# Morphology and Syntax of the Tamil Language

**Kengatharaiyer Sarveswaran**
University of Jaffna. Sri Lanka.
sarves@univ.jfn.ac.lk

## Abstract

This paper provides an overview of the morphology and syntax of the Tamil language, focusing on its contemporary usage. The paper also highlights the complexity and richness of Tamil in terms of its morphological and syntactic features, which will be useful for linguists analysing the language and conducting comparative studies. In addition, the paper will be useful for those developing computational resources for the Tamil language. It is proven as a rule-based morphological analyser cum generator and a computational grammar for Tamil have already been developed based on this paper. To enhance accessibility for a broader audience, the analysis is conducted without relying on any specific grammatical formalism.

## 1  The Tamil language

Tamil is a Southern Dravidian language spoken natively by more than 78 million people across the world,[1] including in India, Sri Lanka, Malaysia, Singapore, South Africa, Mauritius, Fiji, and Burma (Krishnamurti, 2003). It is also one of the top 20 languages in the world based on population count (Simons and Fennig, 2017). Tamil has more than two millennia of continuous and unbroken literary tradition (Hart, 2000) and has been recognised as a classical language by the Government of India. It has the longest literary tradition among all the Dravidian languages (Lehmann, 1998). It is an official language of Sri Lanka and Singapore and has official regional status in the states of Tamil Nadu and Pondicherry in India. It has also been recognised as a minority and indigenous language in several countries, including Malaysia, Mauritius, and South Africa. Tamil is taught in schools in several countries, including Australia, Canada, and the United Kingdom.

The earliest classical Tamil is called *sangattami̱l*. Modern Tamil is known for its diglossic nature whereby the spoken varieties, referred to as the low variety or *ko̱duntami̱l* and the written variety, also referred to as the high variety or *centami̱l* (Britto, 1991;

---
[1] http://www.languagesgulper.com/eng/Tamil.html

Krishnamurti, 2003), differ phonologically and morphologically. The spoken forms of Tamil vary depending on the regions where they are spoken. This is mainly due to language contact and politics (Schiffman, 2008). Within Sri Lanka, there are several spoken varieties of Tamil. At times, one encounters cases where one speaker may not understand the other. Further, no common agreement has been made among different governments (at least between Sri Lanka and Tamil Nadu in India) on the choice of Tamil words for certain technical terms. Consequently, terminological variation is also present. Although there are variations in spoken Tamil, the linguistic structure of written Tamil remains mostly uniform across the different regions.

## 1.1 Tamil script

The Tamil language has its own script, which is nowadays referred to as Tamil script and follows the Abugida or Alphasyllabary writing system where a pure consonant and a vowel are written together as one unit or syllable. For instance, the consonant த் (*t*) together with the vowel உ (*u*) form the single unit, a composite character து (*tu*). The alphabet has 12 vowels, 18 consonants (non-composite characters), and 216 composite characters, which are formed when combining pure consonants and vowels together as a single unit. Tamil also has a special character ஃ (*ah*) — a guttural, which is categorised neither as a vowel nor as a consonant, but is in the alphabet. The University of Madras lexicon describes this letter as "The 13$^{th}$ letter of the Tamil alphabet occurring only after a short initial letter and before a hard consonant". In addition to this total of 247 Tamil letters, some letters from the *Grantha* alphabet, which is the script used to write Sanskrit in South India, are also widely used along with the Tamil alphabet.

Apart from the alphabet, Tamil has its own native numerals, which include whole numbers and fractions, metrics, and symbols for various concepts such as time that are still being documented. It is, however, unfortunate that most current Tamil speakers are not aware of these symbols, as they are not included in the school curriculum, even though some of these symbols are still used in almanacs. However, they were commonly used in books printed up until the early 20$^{th}$ century, with Tamil numerals even being used for page numbering.

## 1.2 Encoding Tamil letters

Apart from the information about how the Tamil script works, understanding how it is encoded in computers using Unicode[2] is important in order to develop language processing tools and to appreciate the complexities involved in processing the Tamil script. Each syllable in Tamil has a single code point. For instance, all the vowels and pure consonants in the Tamil alphabet are each encoded with a single code point. All the composite characters, on the other hand, are encoded with two Unicode points; one to represent the consonant letter, and the other to represent the vowel modifier (also called dependent vowel sign),[3] which is applied to form the composite. For instance, து (*tu*) will have two Unicode points,[4] one for த (*t*) and the other for the ு (*u*)- vowel modifier. Similarly, தௌ (tou) also has two Unicode points corresponding to த (*t*) and ௌ (*ou*) even though it is represented by three glyphs. Typing these vowel modifiers and handling them as part of computer programming is not a straightforward task. Unfortunately, some applications still struggle with rendering these vowel modifiers.

---

[2]https://home.unicode.org/

The Tamil script was first encoded in the Unicode version 1.1 in the year 1993.[5] Since then, several Tamil language processing tools have been built around the world. Prior to the introduction of Unicode, ASCII encoding was used along with special fonts to render Tamil graphemes. Although this is not useful for Tamil language processing, this obsolete approach is sometimes used to create text. This adds an extra level of challenges when processing Tamil texts since these differences are not visible to the naked eye.

## 1.3 Tamil Grammars

The grammars of Tamil may be divided into ones that are composed by native scholars and those that Europeans have written to facilitate the acquisition of the Tamil language, mainly by foreigners (Pope, 1979). *tolkāppiyam* is recognised as the earliest scholarly work on Tamil grammar by a native scholar (George, 2000). The date of this publication is not exact, yet it is believed to have been published more than 2500 years ago. This is considered to be a derived work of an even older work called *agastyam* (Pope, 1979), the whole work of *agastyam* is not extant. The other notable and still widely used and quoted Tamil grammar is called *nannūl*, which is dated back to the 13[th] century. Modern-day Tamil grammar textbooks used in Sri Lanka are themselves based on *nannūl*. Apart from *tolkāppiyam* and *nannūl*, several derived works have been published by native scholars from time to time. An important factor about these works usually accommodates the evolution of the Tamil language. Until the 16th century, these old grammars, dictionaries, treatises on medicine, and literature were written in verse form, making them interpretable only by Tamil pundits or experts — a non-trained person of the current generation would not be able to understand this genre of Tamil. These texts are mostly understood through the commentaries written by scholars. At times, even these commentaries require a further simplified version in order to be understood. For instance, a scholar called *parimēlaḻakar* wrote a commentary (Nachinarkkiniyar, 1937) for the first Tamil grammar work *tolkāp-*

---

[3] https://unicode.org/charts/PDF/U0B80.pdf
[4] https://unicode.org/charts/PDF/U0B80.pdf
[5] https://www.unicode.org/standard/supported.html

*piyam* in the 13<sup>th</sup> century. Later, many simplified commentaries were published based on *parimēlaḻakar*'s work, due to difficulties in understanding the structure and content of those commentaries.

From the 16<sup>th</sup> century onwards, Europeans and others, mostly missionary scholars, started publishing books in Tamil and writing Tamil grammar books. However, since the 19<sup>th</sup> century, no notable complete linguistic studies have been done on Tamil (Annamalai et al., 2014). On the other hand, the language has evolved significantly due to modern subject areas, new kinds of literature, technological advancements, and influences from other languages, particularly as a result of global communication and movement. *eḻuttu* (orthography), *col* (morphology+syntax), *pāeruḷ* (semantics), *yāppu* (prosody), *aṇi* (rhetorical embellishment) constitute the five key parts of Tamil grammar (Shanmugadas, 1982; Nachinarkkiniyar, 1937; Senavaraiyar, 1938). Not all of the grammars cover all of these parts. For instance, the widely used *Nannūl* covers only the first two.

Tamil displays a relatively free constituent order, though it primarily follows a Subject-Object-Verb (SOV) structure in formal writing. Using a corpus-based study, Futrell et al. (2015) show that Tamil has the highest word order freedom when compared to the 100+ languages that are available in the Universal Dependencies treebank collection.

In traditional grammar, Tamil words are categorised into four types, namely nouns, verbs, particles and intensifiers (Senavaraiyar, 1938; Thesikar, 1957). However, modern linguists classify Tamil words into *peyar* (noun), *viṉai* (verb), *peyaraṭai* (adjective), *viṉaiyaṭai* (adverb) and *iṭaiccol* (particle) (Nuhman, 1999; Annamalai et al., 2014). Nominative, accusative, dative, instrumental, sociative, locative, ablative, genitive and vocative are the nine cases that mark nouns as described in the literature (Lehmann, 1993). In addition to the case features, all nouns in Tamil can be categorised into *uyartiṇai* (rational) and *aḵṟiṇai* (irrational). Entities marked as rational are those perceived as being able to think on their own, while the rest are termed as irrational. For instance, animals, trees, furniture items, and infants are considered irrational, while humans and gods are categorised as rational. This (ir)rationality based marking differs from splits in terms of human vs. non-human, or animacy. For instance, infants are considered irrational just as animals or inanimate objects, even though infants are human and animate. An adult human entity can also still be marked as irrational if they behave in an insane manner. This marking is in turn reflected in the noun-verb agreement. Tamil verbs have complex morphosyntactic relations, which take auxiliary items such as question particles, emphatic particles, and conjunctive markers, in addition to regular inflections such as tense, person, gender, number, and honorifics to build a surface form.

# 2 Tamil Morphology

Tamil is an agglutinative language in which grammatical components are suffixed to root words. These suffixes can be identified and separated from the rest of the word through a complex segmentation process to determine the root words. This section provides an overview of the morphology of nouns, verbs, adjectives, and adverbs.

## 2.1 Nominal Morphology

Nouns in Tamil are primarily marked for number and case. These features are expressed as suffixes that are added to a 1. lemma form or 2. oblique form (which will be discussed later in this section). In addition, an oblique form can take euphonic and other phonologically motivated material before the number and case suffix (Caldwell, 1998; Shanmugadas, 1982; Lehmann, 1993).

The number or/and case suffix can be just added to a lemma form, as in example (1) (a) and (b). Apart from such simple suffixation, a glide letter ( ய் / வ்) (y/v) may be inserted, as in Example (1) (c), or a consonant (க்/ ப் / ச் / ற்) (k/p/c/ṟ) may be inserted, as in Example (1) (d). In addition to these, an assimilation process may also occur where an existing letter is changed to another, as in Example (1) (e), where ல் becomes ற்.

(1)     (a)     கதிரைகள் (*katiraikaḷ*) 'chairs'
                   கதிரை   -கள்
                   *katirai*   *-kaḷ*
                   chair     PL

      (b)     கதிரைக்கு (*katiraikku*) 'to a chair'
                   கதிரை   -க்கு
                   *katirai*   *-kku*
                   chair     DAT

      (c)     கதிரையுடன் (*katiraiyudan*) 'with a chair'
                   கதிரை   -ய்    -உடன்
                   *katirai*   *-i*     *-udan*
                   chair     GLIDE   SOC

      (d)     பூக்கள் (*pūkkaḷ*) 'flowers'
                   பூ        -க்     -கள்
                   *pū*       *-k*      *-kaḷ*
                   flower   SANDHI   PL

      (e)     புற்கள் (*puṟkaḷ*) 'grasses'
                   புல்      -கள்
                   *pul*      *-kaḷ*
                   grass    PL

Oblique forms are generated primarily by doubling the last consonant, adding a compulsory oblique suffix, or by deleting last letter of the lemma. This process makes a stem eligible to receive case markers in some noun classes (Krishnamurti, 2003). The compulsory oblique suffixes are referred as *cāriyai* in Tamil grammar books (Nuhman, 1999; Thesikar, 1957). அம் (*am*), அத்து (*attu*), and அற்று (*aṭṭu*) are the widely seen oblique suffixes in texts. Examples for obliques are shown in (3).

Number and case suffixes are bound morphemes. However, in modern Tamil, case suffixes can also be seen separately as free morphemes. Lehmann (1993) also points out the possibility of their occurrence as free morphemes in Tamil.

### 2.1.1 Structure of simple Nouns

The morphology of a simple Tamil noun, without any derivations or compounding, is depicted in the formulas in (2) (a) and (b). In Tamil text, nouns with two euphonic markings can also be found, as shown in (2) (b). Euphonic morphemes such as அன் (*an*) and இன் (*in*) are purely phonological increments (Lehmann, 1993). However, the exact functions of these increments are yet to be fully understood. (3) illustrates various ways a noun can be formed from an oblique base. A stem can take only one oblique suffix, but it can accommodate multiple instances of euphonic morphemes.

(2)     (a) Noun = lemma-form (+plural) (+case)  
        (b) Noun = oblique-form (+plural) (+euphonic)$^2$ (+case)

(3)     (a) மரங்களினால் (*marangkaḷinaal*) 'by trees'

| மரம் | கள் | இன் | ஆல் |
|---|---|---|---|
| *maram* | *kaḷ* | *in* | *aal* |
| tree | PL | EUPH | INST |

        (b) மரத்தினுக்கு (*maratinukku*) 'to a tree'

| மரம் | அத்து | இன் | கு |
|---|---|---|---|
| *maram* | *attu* | *in* | *ku* |
| tree | OBL | EUPH | DAT |

        (c) ஆவினுக்கு (*aavinukku*) 'to a cow'

| ஆ | இன் | உ | கு |
|---|---|---|---|
| *aa* | *in* | *u* | *ku* |
| cow | EUPH | EUPH | DAT |

Nouns in Tamil can also be derived from verbs. However, detailed coverage of derivational morphology is not included in this work, except the formation of verbal nouns, which is discussed in Section 2.2.

### 2.1.2 Plurals in Tamil

Tamil noun stems are singular by default. The plural suffix is a bound morpheme in Tamil that is marked by கள் (*kaḷ*). However, this marker is also used as an honorary marker in present-day Tamil usage, especially when added to the third person pronoun அவர்-கள் (*avar-kaḷ*) 'he.3SEH'.

### 2.1.3 Cases in Tamil

Traditional grammarians have identified 8 cases including a vocative (Senavaraiyar, 1938; Thesikar, 1957). However, modern linguists (Nuhman, 1999; Paramasivam, 2011; Lehmann, 1993) argue that the instrumental case in traditional grammar should be treated as two, namely instrumental case and sociative case. This modern classification along with the respective case markers are shown in Table I.[6]

Some of the case markers in Tamil are free morphemes, specifically the locative, sociative, ablative, and instrumental markers (Lehmann, 1993). For example, சாவியால் (*saaviyaal*) 'key.INST' can also be expressed as சாவி மூலம் (*saavi muulam*). However, these free morphemes are not exclusively used for marking cases. For instance, the lexical meaning of மூலம் (*muulam*) is 'origin,' and it is also employed to mark the instrumental case.

Tamil does not have a definite marker. Instead, definiteness is indicated using demonstrative markers or by marking irrational objects with the accusative case marker. (Lehmann, 1993). While the object of a sentence may be marked with the accusative case, this is compulsory only for rational objects (Lehmann, 1993; Nuhman, 1999). Thus, when an irrational noun carries an accusative marker, it marks that noun as definite, classifying Tamil as a Differential Object Marking (DOM) language.

---

[6]Thesikar (1957) lists 28 case markers for the locative case and 10 case markers for the vocative case. Collectively, most of these markers are rarely used in present-day Tamil.

Table I: Case markers in Tamil

| Morpheme | Morphs / suffixes | Example |
|---|---|---|
| Nominative | - | மரம் (*maram*) 'a tree' |
| Accusative | ஐ (*ai*) | மரத்தை (*marattai*) 'the tree' |
| Instrumental | ஆல் (*aal*) | மரத்தால் (*marattaal*) 'using a tree' |
| Sociative | ஒடு, உடன் (*oṭu, uṭan*) | மரத்துடன் (*marattuṭan*) 'with a tree' |
| Dative | கு,க்கு,அக்கு,உக்கு (*ku,kku,akku,ukku*) | மரத்துக்கு (*marattukku*) 'to a tree' |
| Ablative | இல்/இன்+இலிருந்து (*il/in+iliruntu*) | மரத்திலிருந்து (*marattiliruntu*) 'from a tree' |
| Genitive | அது, உடைய, இன் (*atu, uṭaiya, in*) | மரத்தின் (*marattin*) 'of a tree' |
| Locative | இல், இடம் (*il, iṭam*) | மரத்தில் (*marattil*) 'on a tree' |
| Vocative | ஆ, ஏ, ஈ (*aa,ee,ii*) | மரமே (*marame*) 'oh a tree!' |

### 2.1.4 Nominal paradigm

Rajendran (2009) has proposed a noun paradigm incorporating 26 classes based on their morphophonological properties. Among these 26 classes, nine classes are used to capture the morphophonological rules pertaining to pronouns.

Classes of the noun paradigm that are not pronominal are shown in Table II. One word is selected to represent each class, as Rajendran (2009) does, and the classes are named on the basis of that representative lexical item. Class distinctions are here determined by the last vowel modifier (or vowel) [class 1-3,6-8], consonant [classes 4-5,9-16], or whether it is a two-letter word [class 10 and 11], whether it is a two-letter word and the first letter ends with a long vowel modifier (or vowel)[classes 13 & 14, 6 & 7]. Based on this, a script has also been developed by me to classify nouns into their respective classes.[7] Classes 6 & 7, and 13 & 14 have been separated into different classes even though they have the same last character in the orthography. This distinction has been motivated by the fact that nouns in these classes differ in conjugation patterns. Certain nouns can have several conjugational forms, which are widely accepted to mark plurality or case. Tamil grammar texts also accept such multiple forms, and refer it as போலி (*pōli*) 'fake' (Nuhman, 1999; Thesikar, 1957). For instance, *naal* 'day' shows conjugations of class 4 and class 14 in present-day Tamil. Therefore, *naal* can be included *naal* in both classes.

---
[7]https://github.com/sarves/Tamil-Noun-Classifier

### 2.1.5 Nominal conjugational forms

A total of 36 declension forms are utilised for Tamil nouns, covering plural and case conjugations, along with external *Sandhi* markers. Each noun root is acted upon by case markers in both its singular and plural forms. Furthermore, nouns in their dative or accusative forms can also be influenced by one of four external *Sandhi* markers. In sum, this results in 36 nominal conjugational forms.[8]

## 2.2 Verbal Morphology

Tamil verbs have complex morphosyntactic relations, which take auxiliary items such as question particles, emphatic particles, and conjunctive markers, in addition to regular inflections such as tense, person, gender, number, and honorifics to build a surface form.

The structure of a simple verb in Tamil is shown in Formula (4).[9] The euphonic particle அன் (*an*) is optional and when present, it seems to be used to add a dimension of politeness to the verb in current usage. The medial particle is used to realise tense (past, present and future), or to negate the verb (Pope, 1979; Lehmann, 1993; Paramasivam, 2011).[10] The terminal suffix of a finite verb is used to realise multiple types of information such as number, person, gender, and rationality (or status) (Pope, 1979; Lehmann, 1993). However, this terminal suffix cannot be chunked or divided to extract these information — it is a portmanteau morph. For instance, in (5), -*aan* denotes that the terminal-suffix is third person, singular, masculine, and rational. As for other morphosyntactic features, Tamil has singular and plural values for number, first/second/third person values, and three gender values: masculine, feminine and neuter. In addition to these three genders, a fourth class called 'epicene' is used to mark the third person plural forms of rational entities (Lehmann, 1993). This is what affects the choice of the correct terminal suffix.

(4)     Verb=lemma-form+<medial-particle>+(euphonic-particle)+<terminal-suffix>

In addition to simple verbs, Tamil also has complex or compound verbs with more than one verbal root, which may express mood, aspect, negation, interrogative, emphasis, speaker perspective, conditional, and causal relations (Annamalai et al., 2014). Agesthialingom (1971) claims that Tamil can have up to four verbal roots in one verb form. For instance, there are four verbal roots in the complex verb in (5): *vaa* 'come', *koḷ* 'hold', *iru* 'be' and *iru* 'be'. *koḷ* 'hold' and *iru* 'be' in the middle together signal a continuous aspect. In (5), we observe that verbal conjugation is only expressed on the last verbal root of the sequence, where it takes tenses, person, number, and gender (PNG) marking. The preceding verbs appear either in a participial or infinitival form.

---

[8]Borrowed words from Sanskrit also can take a nagation marker as the prefix. For instance, நியாயம் (niyāyam) 'Justice' becomes அநியாயம் (aniyāyam) 'Injustice'.

[9]There are two types of transitives in Tamil. The first type is indicated by specific markers that distinguish between intransitive and transitive forms, which we can refer to as derived transitives. In the second case, transitivity is inherent to the root itself, and there is no corresponding intransitive form; this can be can refer to these as inherent transitives.(Agesthialingom, 1971) This work does not distinguish derived and inherent transitives.

[10]Old Tamil has two tenses, past and non-past. There are five allomorphs for past tense: *-t-*, *-nt-*, *-in-*, *-i-*, *-tt-*, and three for non-past tense: *-v-*, *-p -*, *-pp-*.

**Table II:** The Tamil nominal paradigm

| No. | Class name | Plural Marker | Sample case markers |
|---|---|---|---|
| 1 | கடா (*kaṭaa*) 'male goat' or பசு (*pasu*) 'cow' | க்-கள் (*k-kaḷ*) | வை, வால், வுடன், வுக்கு (*vai, vaal, vuṭn, vukku*) |
| 2 | எலி (*eli*) 'rat' or நெய் (*ney*) 'ghee' | கள் (*kaḷ*) | யை, யால், யுடன், யுக்கு (*yai, yaal, yuṭn, yukku*) |
| 3 | ஈ (*ii*) 'house fly' | க்கள் (*kkaḷ*) | யை, யால், யுடன், யுக்கு (*yai, yaal, yuṭn, yukku*) |
| 4 | நாள் (*naal*) 'day' or கால் (*kaal*) 'leg' | கள் (*kaḷ*) | லை, ஆ, உ, ஓ (*i, aa, u, oo*) |
| 5 | பலர் (*palar*) 'many people' | - | லை, ஆ, உ, ஓ (*i, aa, u, oo*) |
| 6 | காடு (*kaaṭu*) 'forest' | கள்(*kaḷ*) | ட்டை, ட்டால், ட்டுக்கு, ட்டோடு (*ṭai, ṭaal, ṭukku, ṭooṭu*) |
| 7 | வண்டு (*vandu*) 'beetle' | கள் (*kaḷ*) | லை, ஆ, உ, ஓ (*i, aa, uu, oo*) |
| 8 | ஆறு (*aaru*) 'river' | கள் (*kaḷ*) | றை, றால், க்கு, றோடு (*ṟai, ṟaal, kku, ṟooṭu*) |
| 9 | கண் (*kan*) 'eye' | கள் (*kaḷ*) | ணை, ணால், ணுடன், ணுக்கு (*ṇai, ṇaal, ṇuṭan, ṇukku*) |
| 10 | பொன் (*pon*) 'gold' | கள் (*kaḷ*) | னை, னால், னுடன், னுக்கு (*nai, naal, nuṭan, nukku*) |
| 11 | மாணவன் (*maanavan*) 'student' | ர்கள் (*rḷ*) | னை, னால், னுடன், னுக்கு (*nai, naal, nuṭan, nukku*) |
| 12 | புல் (*pul*) 'grass' | ற்கள் (*ṟkaḷ*) | லை, லால், லுடன், லுக்கு (*lai, laal, luṭan, lukku*) |
| 13 | முள் (*mul*) 'thorn' | ட்கள் (*ṭkaḷ*) | ளை, ளால், ளுடன், ளுக்கு (*ḷai, ḷaal, ḷuṭan, ḷukku*) |
| 14 | நாள் (*naal*) 'day' | ட்கள் (*ṭkaḷ*) | லை, ஆ, உ, ஓ (*i, aa, uu, oo*) |
| 15 | மரம் (*maram*) 'tree' | ங்கள் (*ṅkaḷ*) | த்தை, த்தால், த்துக்கு, த்தோடு (*ttai, ttaal, ttukku, ttooṭu*) |
| 16 | சுவர் (*suvar*) 'wall' | கள் (*kaḷ*) | ற்றை, ற்றால், ற்றுக்கு, ற்றோடு (*ṟṟai, ṟṟaal, ṟṟukku, ṟṟooṭu*) |

(5) வந்துகொண்டிருந்திருக்கிறான்
*vantukoṇṭiruntirukkiraan*
*vantu-koṇṭiru-iru-kkir-aan*
come.VPART-hold_be.VPART-be-PRES-3SMR
'(he) has been coming'

Complex verbs in Tamil can be written as separate tokens, as in (6), or as a single token, as in (7).

(6) வாங்கச் செய்தான்
*vang-a-c sei-t-aan*
buy-INF-SANDHI_C do-PAST-3SMR
'(he) made someone buy.'

(7) வாங்கச்செய்தான்
*vang-a-c-sei-t-aan*
buy-INF-SANDHI_C-do-PAST-3SMR
'(he) made someone buy.'

(8) வாங்கிக்கொடுத்தான்
*vāṅkik-koṭu-tt-āṉ*
buy-VPART-SANDHI_K-give-PAST-3SMR
'(he) bought for someone'

The set of verbs that form complex verb forms in conjunction with the main verb are identified and categorised based on their structure and function, primarily relying on the discussions in Boologarambai (1986). However, further research is necessary to identify additional complex verbal conjugational forms and their respective functions.

**Table III:** List of complex verb forms

| Verb types | Verb roots[11] | Structure of a finite verb |
|---|---|---|
| Aspectual verbs | இரு (*iru*) 'be', கொண்டிரு (*koṇṭiru*) 'keep', விடு (*viṭu*) 'leave', முடி (*muṭi*) 'finish' | main-verb+VP +aspectual-verb+TENSE+PNG |
| Attitude verbs | போ (*po*) 'go', போடு (*poṭu*) 'put', தள்ளு (*taḷḷu*) 'push', தீர் (*tiir*) 'solve', தொலை (*tolai*) 'get lost' | main-verb+VP +attitude-verb+TENSE+PNG |
| Non-attitude or Light Verbs | இடு (*iṭu*) 'put', கொடு (*koṭu*) 'give', பார் (*paar*) 'see', வா (*vaa*) 'come' | main-verb+INF +non-attitude-verb+TENSE+PNG |
| Modal verbs | வேண்டு (*veeṇṭu*) 'want', கூடு (*kooṭu*) 'may' | main-verb+INF +modal-verb+TENSE+PNG |
| Causatives | பண்ணு (*paṇṇu*) 'make', செய் (*sei*) 'make', வை (*vai*) 'cause' | main-verb+INF +causative-verb+TENSE+PNG |
| Passivisers | படு (*paṭu*) 'suffer', பெறு (*peru*) 'get' | main-verb+INF +passiviser+TENSE+PNG |

### 2.2.1 Verbal paradigm

Tamil verbs can be classified on the basis of criteria that can be either morphological, syntactic or semantic (Paramasivam, 2011). Many scholars, including Lisker (1951), Graul (1855), and Arden (1910) have classified verbs on the basis of what their morphophonemic changes they display as part of their conjugations. Graul (1855) has provided an early classification on which other scholars have built their proposals, including Irākavaiyaṅkār (1958) and Sithiraputhiran (2004). Graul's classification has also been adopted for the Tamil lexicon project (Rajaram, 1986). This classification of Tamil verbal lemmas includes 12 categories or classes and is based on the tense markers as displayed on the verbs. As shown in Table IV, when verb forms are conjugated for tense markers, these are not just added to the lemma form via concatenation. Rather changes or new letters may be introduced, and these constructions are handled with the use of alternation rules. For instance, in class 12, when the past tense marker is coined, it is not just added to the lemma, e.g. (நட+த் (naṭa+t) 'walk.SANDHI_T. Instead, the letter ந் (nt) is inserted as in நட+ந்+த் (naṭa+n+t). A similar behaviour presents in other classes as well.

---

[4]Note: I have given the literal meaning of the verbs in the column - Verb roots. However, when functioning as an auxiliary or light verb, these meanings may not hold.

In addition to these classes, I have identified five irregular verbs that do not fit into the paradigm given in Table IV: காண் (kaaṇ) 'see', வா (vaa) 'come', சா (saa) 'die', தா (taa) 'give', வே (vee) 'boil'.

**Table IV:** The Tamil Verbal

| No. | Class name | Past tense marker | Present tense marker | Future tense marker |
|---|---|---|---|---|
| 1 | செய் (sei) | த் (t) | கிற், கின்ற் (kir, kinr) | வ், உம் (v, um) |
| 2 | ஆள் (aaḷ) | ட் (ṭ) | கிற், கின்ற் (kir, kinr) | வ், உம் (v, um) |
| 3 | கொல் (kol) | ற் (ṟ) | கிற், கின்ற் (kir, kinr) | வ், உம் (v, um) |
| 4 | கடி (kaṭi) | த் (t) | கிற், கின்ற் (kir, kinr) | வ், உம் (v, um) |
| 5 | அஞ்சு (angu) | இன் (in) | கிற், கின்ற் (kir, kinr) | வ், உம் (v, um) |
| 6.1 | அடு (aṭu) | ட் (ṭ) | கிற், கின்ற் (kir, kinr) | வ், உம் (v, um) |
| 6.2 | நகு (naku) | க் (k) | கிற், கின்ற் (kir, kinr) | வ், உம் (v, um) |
| 6.3 | உறு (uru) | ற் (ṟ) | கிற், கின்ற் (kir, kinr) | வ், உம் (v, um) |
| 7 | உண் (uṇ) | ட் (ṭ) | கிற், கின்ற் (kir, kinr) | ப், உம் (p, um) |
| 8 | தின் (tin) | ற் (ṟ) | கிற், கின்ற் (kir, kinr) | ப், உம் (p, um) |
| 9 | கொள் (koḷ) | ட் (ṭ) | கிற், கின்ற் (kir, kinr) | ப், உம் (p, um) |
| 10 | நில் (nil) | ற் (ṟ) | கிற், கின்ற் (kir, kinr) | ப், உம் (p, um) |
| 11 | அபகரி (abakari) | த் (t) | க்கிற், க்கின்ற் (kkir, kkinr) | ப்ப், உம் (pp, um) |
| 12 | நட (naṭa) | த் (t) | க்கிற், க்கின்ற் (kkir, kkinr) | ப்ப், உம் (pp, um) |

### 2.2.2 Verbal conjugational forms

Annamalai et al. (2014) have identified 254 forms for each Tamil verb following a rigorous analysis of their corpus of contemporary texts. Some verbs may, however, not take all of the 254 forms. Rajaram (1986) has identified 21 forms for each verb from a pedagogical perspective. On the other hand, Kumar et al. (2010) claims that a Tamil verb lemma can take up to 8,000 forms if derivations are also considered. Sarveswaran et al. (2021) has compiled over 3,000 verb lemmas and identified close to 600 forms for each verb lemma. For example, 3.13 shows 582 forms for the verb lemma நட *naṭa* 'walk', including the suffixation of *Sandhi*. This count excludes the causative derivation lemma நடத்து *naṭattu* 'make walk/conduct'. Additionally, this list does not account for the suffixation of auxiliary verbs, including modal or aspect aspects. When considering auxiliary verbs as well, each verb can potentially yield several thousand forms. Sarveswaran et al. (2021) has made efforts to create these compounds, compiling over 57 million of them for all 3,000+ verb lemmas.[12] While some of these words may not be in current use or not used such formations remain possible.

## 2.3 Adjectival Morphology

In Tamil, everything which precedes and modify a noun is called பெயரடை *peyaraṭai*. This includes adjectives, nominal modifiers, demonstrative pronouns, numeric modifiers, and adjectival participles. There are two sorts of adjectives in Tamil: pure adjectives and derived adjectives. Pure adjectives are words that are used as adjectives without requiring any suffixation. The attributive type of adjectival derivation, which is formed by adding the suffix ஆன *-aana* 'an adjectiviser' to nouns, as shown in (9). This suffix is derived from a verb ஆகு *-aaku* 'become'. உள்ள *-uḷḷa* 'in' (rough translation) is the other adjectiviser used in Tamil (Nuhman, 1999).

(9) உயரமான
*uyaram-aana malai*
tall-be mountain
'tall mountain'

Example in (10) shows an instance of adjectival modification, where the verb come.PAST take an adjectival suffix *a* to modify the noun. This follows the structure of relative clause. However, instead of a relative pronoun, here we have a noun. Further, in relative clause, we mark *a* as a relativiser.

Example in (11) shows a nominal modifier. Similarly, we can also have numeric modifiers precedes to nouns.

(10) வந்த        பையன்
*vaa.nt.a     paiyan̲*
come.PAST.ADJ boy
'The boy who came'

(11) அமைதிப்    படை
*amaiti-p    paṭai*
peace.SANDHI_P force
'A peace force'

Tamil has three demonstratives to indicate spatial deixis: இந்த *inta* 'this', அந்த *anta* 'that', and உந்த *unta* 'between this and that', which are marked by the respective demonstrative markers in Tamil: இ *i*, அ *a*, and உ *u*, respectively. Among these markers,

---
[12]https://www.kaggle.com/datasets/sarves/tamilverbs

*uṉṭa* is primarily used in the northern region of Sri Lanka — Jaffna, and its usage has diminished in other parts of the world.[13]

Adjectives can be reduplicated to modify plural nouns as shown in Example (12), which is taken from (Nuhman, 1999). Further, superlative forms are derived by addition modifiers to adjectives in Tamil, as shown in (13). However, in poetic writing this can even be preceded by மிக *mika* 'very' to mark much more tininess.

(12) சிறிய சிறிய வீடுகள்
*ciṟiya ciṟiya  vīṭukaḷ*
small small   houses
'small small houses'

(13) சின்னஞ்சிறிய வீடு
*ciṉṉañ-ciṟiya  vīṭu*
small-small   house
'The smallest house'

## 2.4 Morphology of adverbs

Everything which precedes and modify a verb is called வினையடை *viṉaiyaṭai*, in Tamil. Similar to adjectives, adverbs in Tamil come in two types: pure adverbs and derived adverbs. Pure adverbs are word forms that are themselves used as adverbs without requiring any sort of suffixation; for instance, in (14), the adverb *nēṟṟu* 'yesterday' does not have any suffixation on it. Several adverbs are subsumed under this type, namely, temporal, frequency, place, degree, and affirmation (Nuhman, 1999). Manner adverbs fall under the second type, i.e. derived adverbs. These derived adverbs are formed by adding an adverbial suffix ஆக/ஆய் *-āka/āy* 'become' to nouns, as shown in (15).

(14) நேற்று வந்தான்
*nēṟṟu  vantāṉ*
yesterday came (he)
'He came yesterday'

(15) வேகமாக வந்தான்
*veham-āka  vantāṉ*
speed-ADV  came (he)
'(He) came fast'

In this section, the morphology of the Tamil language has been discussed, particularly focusing on the inflectional morphology of nouns, verbs, adjectives, and adverbs. Brief mention has also been made of derivational morphology where relevant. It is now evident

---

[13]Ronald E. Asher, in the preface of Suseendirarajah (1999) claims that knowledge of Tamil, specifically Jaffna Tamil, a variety spoken in Sri Lanka, is a vital element that forms part of the understanding of the Dravidian family of languages. This is particularly so since certain features of the language have been preserved in Jaffna Tamil that have been lost elsewhere.

that Tamil exhibits a complex morphological system.

## 3 Syntax of Tamil

Over the years, Tamil has been in contact with several other languages. It is no longer spoken in some of the countries in which it was spoken several decades ago. Further, due to the migration that has been happening, Tamil is now spoken all over the world and has been in contact with a number of Western languages as well. Apart from this array of language contact, the grammaticalisation of various words in Tamil through time makes the study more complex. There are, however, only a small number of works that relate to modern Tamil grammar. Similarly, there is no comprehensive grammar for modern Tamil written after the 13<sup>th</sup> century *naṉṉūl* grammar. It is the derived grammar that is based on *naṉṉūl* (Thesikar, 1957) that is used as a high school Tamil textbook (Nuhman, 1999). Tamil requires a significant amount of proper linguistic study to understand and describe its grammar, especially syntax of modern Tamil. There has been an attempt to develop a modern grammar for Tamil by several scholars in Tamil Nadu, including E. Annamalai. However, no parts of the grammar have been released yet.

This section highlights certain syntactic structures of Tamil extracted from a corpus. Furthermore, various syntactic analyses in grammar books and research papers have been followed, although they also require further linguistic exploration. The use of modern linguistic theories has shed some light on a few constructions such as complex predicates and light verbs, as detailed in Sarveswaran and Butt (2019), as a better understanding of these is essential for a complete understanding of their structure and semantics.

Tamil is a highly free word order language, although Subject-Object-Verb (SOV) is the commonly used constituent order. For this reason, and for further simplification, variants other than SOV are not covered in this overview.

### 3.1 Nouns and their structures

Nouns in Tamil are marked for number and case. The number values available are singular and plural. Case[14] in Tamil has received much attention but has evolved over time. Modern Tamil grammarians propose nine cases; some are discussed in detail below. There is evidence that some case forms in Tamil are also used to mark other cases. For instance, the accusative case can sometimes be used to express the instrumental (Joseph, 1893). This section will not make any further reference to this additional syncretic complexity. These aspects require further linguistic exploration.

---

[14] The analysis of whether the given case marker is a 'true' case marker or 'just' a postposition has not been conducted. The approach has been to follow the existing literature, although there are instances where further analysis may be needed. Nonetheless, this issue has been a longstanding concern in the Tamil case system (Schiffman, 2004; Caldwell, 1998). In fact, Caldwell (1998) notes the following regarding the Dravidian case system: "All case-relations are expressed by means of postpositions, or postpositional suffixes. In reality, most of the postpositions are separate words; in all the Dravidian dialects, they retain traces of their original character as auxiliary nouns. Several case signs, especially in the more cultivated dialects, have lost the faculty of separate existence and can only be treated now as case terminations. However, there is no reason to doubt that they are all postpositional nouns originally."

Apart from numbers and case, functional elements like adverbials, postpositions, particles, and clitics can also be attached to nouns. Adverbials are primarily of a spatial or temporal nature. While clitics can be added to nouns irrespective of their number and case marking, particles can be attached to nouns if they are in the nominative, accusative or dative case. Adverbials and postpositions are attached to nouns if they are in the nominative or the dative case.

### 3.1.1 Nominative case

Syntactically unmarked, bare nouns are understood to be nominative-marked. These may function as a subject, predicate, subject-complement, objective-complement, or object in Tamil. Except for dative subjects, subjects in nominative cases usually necessitate subject-verb agreement, where the verb shows agreement with the nominative case marked nominal in person, number, gender (and rationality).

(16) கண்ணன் ஒரு மாணவன்
kaṇṇaṉ oru māṇavaṉ
Kannan.NOM a student.NOM.MASC
'Kannan is a student.'

(17) குமார் தலைவன் ஆனான்
kumār talaivaṉ āṉāṉ
Kumar.NOM leader.NOM.MASC become.PAST.3SMR
'Kumar became a leader.'

Example (16) is a nominal predicative construction — an equative construction. Such constructions are commonly used in Tamil, where the linking verb is dropped in the present tense constructions. In such null-copula constructions, order of phrases play a role and determine the subject. However, for other tenses, a copula *āna* 'become' is introduced to carry tenses, as in (17). The other variant of this is shown in (18) where *āka* — an adverbial suffix — is added to the nominal predicate in (17). Although examples (17) and (18) following are structurally different, quantifying the difference has proven challenging; it necessitates linguistic exploration. In (18), 'āka' is glossed as 'adv' because it serves as an adverbial suffix used to convert nouns into adverbs.

(18) குமார் தலைவனாக ஆனான்
kumār talaivaṉāka āṉāṉ
Kumar.NOM leader.MASC.ADV become.PAST.3SMR
'Kumar became the leader.'

(19) 
| குமார் | இராமனைத் | தலைவன் | ஆக்கினான் |
|---|---|---|---|
| kumār | irāmanait | talaivan | ākkinān |
| Kumar.NOM | Raman.ACC.SANDHI-T | leader.NOM | make.PAST.3SMR |

'Kumar made Raman a leader.'

In N+V predication – light verb construction – N is always in the nominative case and serve as the semantic head of the phrase. For instance, (18) is a resultative structure with an N+V predicate, where N is in the nominative case.

### 3.1.2 Accusative case

The case marker -*ai* is used to mark the accusative case in Tamil. Like other case markers, this is also added to the nominal stem, oblique stem,[15] or a pluralised noun.

Tamil has differential object marking, where objects are marked with and without an accusative case on the basis of rationality. It is obligatory to have the accusative case marking for rational entities as shown in (19). In his work, (Lehmann, 1993) notes that there are a few exceptional instances where rational entities do not take the accusative case marker when functioning as objects. However, no such instances were found in our corpora.

The accusative marking is optional for irrational entities as shown in (20), and adding an accusative case marker onto these irrational entities ends up denoting definitiveness as in (21). However, if the accusative case is not overtly marked, as in (20), the order of the object becomes restricted in relation to other constituents. Basically, the object has to follow the subject; otherwise, the sentence becomes ambiguous.

(20)
| குமார் | பந்து | அடித்தான் |
|---|---|---|
| kumār | pantu | aṭittān |
| Kumar.NOM | ball.NOM | hit.PAST.3SMR |

'Kumar hit a ball.'

(21)
| குமார் | பந்தை | அடித்தான் |
|---|---|---|
| kumār | pantai | aṭittān |
| Kumar.NOM | ball.ACC | hit.PAST.3SMR |

'Kumar hit the ball.'

### 3.1.3 Dative case

The dative case marked by உக்கு -*ukku* has many different functions in Tamil; Lehmann (1993) lists nine different functions. However, the following four different instances of dative marking were found in the corpus:

**1. Indirect objects**

---

[15]Some nouns take a suffix called oblique suffix, which allows the root to take other suffixes.

Ditransitive verbs in Tamil take indirect objects marked with dative. Unlike the accusative case, the case suffix is always added irrespective of rationality.

(22) 
| இது | அரசுக்கு | ஒரு | சவால் | அல்ல |
|---|---|---|---|---|
| itu | aracukku | oru | cavāl | alla |
| this | state.DAT | a | challenge.NOM | not |

'This is not a challenge to the state.'

**2. Benefactive case**

On top of *-ukku*, a suffix *-āka* is added to mark the benefactive case in Tamil. Although this can be considered a separate construction type, since this is somewhat related to the dative marking this is covered under the dative.

(23)
| நாம் | நாட்டுக்காக | சேவை | செய்யும் | குழு |
|---|---|---|---|---|
| nām | nāṭṭukkāka | cēvai | ceyyum | kuḻu |
| we.NOM | country.BEN | service.NOM | do.FUT.REL | group.NOM |

'We are a group that serves the country.'

**3. Dative subjects**

While some scholars consider dative subjects as indirect objects, others, such as Mohanan and Mohanan (1990), and Pappuswamy (2005), accept the concept of dative subjects. This phenomenon is found in other languages as well (Butt et al., 2006). Unlike nominative subjects, dative subjects show mixed behaviours. They, for instance, do not show all the subject properties shown by a nominative subject, including the person-number-gender agreement with the verb.

Verbal predicates with dative subjects marked for third-person, neuter-gender, and singular — are also referred to as default agreement — irrespective of subject's number, gender, rationality, and person. This default agreement is marked using உம் *-um*, as shown in (24). In this construction, the object does not carry the accusative marking. However, from other constructions it is evident that objects can carry overt accusative marking.

(24)
| எனக்குச் | சிங்களம் | தெரியும் |
|---|---|---|
| enakkuc | ciṅkaḷam | teriy-um |
| I.DAT.SANDHI_C | Sinhala.NOM | know.FUT-3SN |

'I know Sinhala.'

Nominal predicates, which represent feelings, sensations, or states of being, with dative subjects also widely used in Tamil. For instance, (25) shows a nominal predicate example with a dative subject.

(25) எனக்குப் பசி
enakku-p paci
I.DAT-SANDHI_P hungry.NOM
'I am hungry'

**4. Goal marker**

Case marker *-ukku* is also used to mark the goal argument of motion verbs in Tamil, as shown in (26). This is analysed as an oblique argument in grammatical framework like the Universal Dependencies and Lexical Functional Grammar.

(26) அண்ணா கண்டிக்குச் சென்றான்
aṇṇā kaṇṭikkuc ceṉṟāṉ
elder-brother.NOM kandy.DAT.SANDHI_C go.PAST.3SMR
'Elder brother went to Kandy.'

### 3.1.4 Instrumental case

The instrumental case is marked by *-āl* to show instrumentation as in (27). In addition, the instrumentation can also be marked via the use of the postposition மூலம் *mūlam* 'with which', which may or may not be suffixed to the root. In other context, *mūlam* also means 'a source'.

Apart from this regular use-case, the instrumental case is also used to mark the agent in passive constructions, as shown in (28), and to form conditional verbal clauses, as discussed later in this chapter.

(27) பையன் சாவியால் கதவைத் திறந்தான்
paiyaṉ cāviyāl katavait tiṟantāṉ
boy.NOM key.INS door.ACC.SANDHI_T open.PAST.3SMR
'A boy opened the door with a key.'

(28) அமைச்சரவையால் புதிய தீர்மானங்கள் முன்னெடுக்கப்பட்டன
amaiccaravaiyāl putiya tīrmāṉaṅkaḷ muṉṉeṭukkappaṭṭaṉa
cabinet.INS new resolutions.NOM forward-take.PASS.PAST.3PLN
'New resolutions were taken-forward by the Cabinet.'

### 3.1.5 Locative case

In modern Tamil, the *-il* marker is used to mark the location in space and time, and mode on irrational and rational nouns (Lehmann, 1993). For instance, (29) shows how *-il* is used to mark the location in space. *-iṭam* is another marker used to mark locative cases only on rational nouns. Lehmann (1993) claims that *-iṭam* expresses the goal of motion, transaction source, emotion target, and temporary possession. For instance, my corpora consist (30), which is an example of a goal of motion of speech.

|     | குமார் | கூட்டத்தில் | பேசினான் |
| --- | --- | --- | --- |
| (29) | kumār | kūṭṭattil | pēciṉāṉ |
|     | Kumar.NOM | meeting.LOC | speak.PAST.3SMR |
|     | 'Kumar spoke at the meeting.' | | |

|     | குமார் | கண்ணனிடம் | பேசினான் |
| --- | --- | --- | --- |
| (30) | kumār | kaṇṇaṉiṭam | pēciṉāṉ |
|     | Kumar.NOM | Kannan.LOC | speak.PAST.3SMR |
|     | 'Kumar spoke to Kannan.' | | |

### 3.1.6 Ablative case

*-iruntu* is used to mark ablative cases on top of locative-cased nouns. For instance, *-il + -iruntu = -iliruntu* would function as ablative case marker for irrational nouns, as in (31). Similarly, *-iṭam + -iruntu = -iṭamiruntu* in turn functions as the marker for rational nouns.

|     | நான் | கொழும்பிலிருந்து | வீட்டிற்கு | வந்தேன் |
| --- | --- | --- | --- | --- |
| (31) | nāṉ | koḻump-il-iruntu | vīṭṭiṟku | vantēṉ |
|     | I.NOM | Colombo.LOC.ABL | home.DAT | come.PAST.1SR |
|     | 'I came home from Colombo.' | | | |

### 3.1.7 Genitive case

உடைய *-uṭaiya*, அது *-atu*, இன் *-iṉ* are the markers used to mark genitive case in Tamil, which is available in the context of possession, and to show a thing's source or a characteristic/trait of something. There is no exact rule to say which marker needs to be used in what context (Schiffman, 2004). (32) shows how *-iṉ* is used to mark possessive. *-atu* is a confusing maker, which is also used to mark third-person, singular, and neuter in verbs and serves as a pronoun to mark the same. In addition, *-atu* functions differently in relative clause constructions. A genitive case example using *-atu* is shown in (33), as in this example, words can be shuffled when *-atu* is used to mark the genitive case. However, it is not possible with other genitive markers.

|     | நான் | அவரின் | பிள்ளை |
| --- | --- | --- | --- |
| (32) | nāṉ | avariṉ | piḷḷai |
|     | I.NOM | he-HON.GEN | child.NOM |
|     | 'I am his child.' | | |

|     | பள்ளிக்கூடம் | குமாரது |
| --- | --- | --- |
| (33) | paḷḷikkūṭam | kumāratu |
|     | school.NOM | Kumar.GEN |
|     | 'Kumar's school.' | |

Further, it is very common in Tamil that these genitive markers are dropped if it does not lead to any ambiguity. For instance, see (34), which is comparable to (32), where genitive case marker *-in* is dropped, yet it shows the possessiveness.

(34) நான் அவர் பிள்ளை
nāṉ avar piḷḷai
I.NOM he-HON.NOM child.NOM
'I am his child.'

### 3.1.8 Sociative case

ஒடு *-oṭu* and உடன் *-uṭaṉ* are the markers used to mark sociative cases in Tamil, see (35). When multiple entities function together as a subject, this conjunction will be reflected in subject-verb agreement, which is discussed later under topic of Coordination (Section 3.7).

(35) நான் அப்பாவுடன் போனேன்
nāṉ appāvuṭaṉ pōṉēṉ
I father.SOC go.PAST.1S
'I went with father.'

## 3.2 Agreement

A nominal subject and verbal predicate agree for rationality, gender, number, and person in Tamil. Tamil grammar textbooks, e.g. Nuhman (1999), claim that there are five gender values, as given below. Although this classification is referred to as gender, the markers do not express gender values alone. Rather it is gender + number + rationality values that are expressed.

1. āṇpāl - masculine singular
2. peṇpāl - feminine singular
3. palarpāl - rational plural
4. oṉṟaṉpāl - irrational singular
5. palaviṉpāl - irrational plural

Four gender values can also be employed - masculine, feminine, epicene,[16] and neuter, without delving into the added complexity of combining GENDER and NUMBER markings being fused onto the same form.

Plurality agreement, although it is obligatory, not found in some of the instances in the corpus for irrational nouns. It is unclear whether this discrepancy may be related to regional differences. For example, (36) was found in the corpus — this may be from

---

[16]Epicene is used to denote singular-rational entities that are not classified as either masculine or feminine. (Lehmann, 1998)

Indian Tamil, although the alternative structure otherwise available in the Sri Lankan context is the one in (37).

(36) மூன்று நாய்கள் வந்தது
mūṉru nāykaḷ vantatu
three dog.PL come.PAST.3SN
'Three dogs came.'

(37) மூன்று நாய்கள் வந்தன
mūṉru nāykaḷ vantana
three dog.PL come.PAST.3PLN
'Three dogs came.'

When it comes to PERSON, there are three values, namely, first, second, and third, as in most other languages. However, Tamil has one more value that is used to mark a deictic reference between the second and the third person. Although special pronouns are used to address this medial deictic, this does not influence any syntactic functions, at least in the corpora used.

In addition to the three main person values, Lehmann (1993) has proposed the fourth person. According to him, this fourth person pronoun[17] — *tāṉ* — is always co-referential with the subject of the same clause or higher. In addition, there is no special agreement displayed between this fourth-person pronoun and the verb. On the other hand, this *-tāṉ* can easily be confused with the *-tāṉ* clitic that is an emphatic marker in Tamil. Based on the data, *tāṉ* is treated as a clitic when it is written together with the headword, as in (38). Otherwise, when *tāṉ* is written as a separate word, it functions as a fourth person pronoun.

(38) நான் செய்தது தவறுதான்
nāṉ ceytatu tavaṟutāṉ
I.NOM do.PAST.REL wrong.NOM.EMPH
'What I did was wrong.'

(39) கண்ணன் தான் பரீட்சையில் சித்தியடையமாட்டான்
kaṇṇaṉ tāṉ parīṭcaiyil cittiyaṭaiya.māṭṭ.āṉ
Kannan.NOM himself exam.LOC pass.will-not.3SMR

என்று நினைத்தான்
eṉru niṉaittāṉ
that think.PAST.3SMR
'Kannan just thought he would not pass the exam.'

In Tamil, although there is a specific suffix with which to mark honorifics on verbal predicates, it is common to use the plural of suffix as a means to mark honorifics. Therefore, agreement for number may not always hold between the subject and the verbal

---
[17]Although I believe that *tāṉ* is a reflexive pronoun, I have used the terminology of Lehmann (1993)

predicate when it comes to honorifics, as shown in (41).

(40)  நான்     உன்னைப்          போகவிடமாட்டேன்
      nāṉ      uṉṉaip           pōka.viṭa.māṭṭ.ēṉ
      I        you.ACC.SANDHI_P go.let.will-not.1S
      'I will not let you go.'

(41)  ஜனாதிபதி        கூட்டத்தில்     பேசினார்கள்
      jaṉātipati      kūṭṭattil       pēciṉārkaḷ
      president.NOM   meeting.LOC     speak.PAST.3PLER
      'The President spoke at the meeting.'

## 3.3 Negation

Negation can be realised morphologically using the *-ā* or *māṭṭu* morphs, or by syntactically using the word *illai*. *-ā* is used to negative all types of verbal constructions, including finite and conditionals, as shown in (42). *-aa* can easily be confused with the question particle. *mattu* is specified as a future negative marker, as shown in (43) (Nuhman, 1999; Lehmann, 1993).

(42)  நீ    வராமலிருந்தால்   அப்பா         கோபிப்பார்
      nī    varāmaliruntāl   appā          kōpippār
      you   come-not-if      father.NOM    get-angry.FUT.3SER
      'Dad will get angry if you don't come.'

(43)  நான்     உன்னைப்          போகவிடமாட்டேன்
      nāṉ      uṉṉaip           pōka.viṭa.māṭṭ.ēṉ
      I        you.ACC.SANDHI_P go.let.will-not.1S
      'I will not let you go.'

The word *illai* can function as a habitual (44), existential (45), and possessive (46) negator — as shown in the examples.

(44)  நான்   அதிகாலையில்    எழும்புவது   இல்லை
      nāṉ    atikālaiyil    eḻumpuvatu   illai
      I      morning.LOC    get-up       no
      'I do not wake up early.'

(45)  அவனுக்குப்         புத்தி    இல்லை
      avaṉukkup          putti     illai
      he.DAT.SANDHI_P    wit.NOM   no
      'He has no wit.'

(46)  எனக்கு    சந்தேகம்      இல்லை
      eṉakku    cantēkam      illai
      I.DAT     doubt.NOM     no
      'I have no doubt.'

## 3.4 Postpositions in Tamil

Postpositions are sometimes referred to as particles (Arden, 1910), and are the equivalent to prepositions in English (Schiffman, 2004), except for their varied syntactic position, since postpositions are suffixed onto the words which they govern. However, analysing and listing all postpositions in Tamil is difficult since most of the postpositions are often nouns or verbs in their origin. Almost any verb in the language can be advanced to candidacy as a postposition (Schiffman, 2004). For instance, *pōṭu* 'put' and *koḷ* 'hold' can be used as postpositions with which to mark instrumental case (Schiffman, 2004) in modern usage. For example, (47) shows how *koḷ* 'hold' is used as a postposition to mark the instrumental case. However, if one can look deeper, it is clear that *koḷ* still functions as a verb and give the object case to stick. Otherwise, there are no ways to explain the case marking on the stick. Therefore, postpositions in Tamil have not been discussed in this version as they require further linguistic exploration. This will be explored in future revisions, step by step.

(47)
| பொலீஸ் | தடியைக்கொண்டு | கூட்டத்தை | விரட்டியது |
|---|---|---|---|
| polīs | taṭiyaik-koṇṭu | kūṭṭattai | viraṭṭiyatu |
| police | stick.ACC-using | crowd.ACC | chase.PAST.3SN |

'Police chased the crowd with stick.'

## 3.5 Verbs and their syntactic structures

Tamil verbs have been mostly analysed from a prescriptive perspective, and most of these studies are based on the very first Tamil grammar called *tolkāppiyam* and a derived piece of work, the *naṉṉūl* (Thesikar, 1957), published in the 13th century CE. From the 18[th] century CE onwards, Western scholars have also contributed to the study of Tamil grammar. However, except for the attempt by Annamalai (2013), none of the scholars has clearly articulated the differences between complex predicates, serial verb constructions (Steever, 2005; Fedson, 1981), complex verbs (Agesthialingom, 1971), and compound verbs (Agesthialingom, 1971; Nuhman, 1999; Fedson, 1981; Paramasivam, 2011).

### 3.5.1 Complex Predicates

The study of complex predicates (CP) has received a great deal of attention in the linguistic literature and a number of distinct interpretations. This research is based on the definition proposed in Butt (1995), which views CPs as being formed when two or more predicational units enter into a relationship of co-predication. Each predicational unit adds arguments to a mono-clausal prediction; a similar definition or idea can also be found in Mohanan (1994) and Alsina et al. (1997). It is important to identify CPs to understand the differences with respect to other potentially confusing categories for the development of computational resources such as computational grammars (Butt and King, 2002), WordNet (Chakrabarti et al., 2007), and machine translation (Kaplan and Wedekind, 1993; Butt, 1994).

Complex predicates are very common in Tamil (Annamalai, 2013). For instance, verbs like வை (vay) 'place', விடு (vidu) 'let go', பார் (paar) 'see/look' may function both as main/full and light verbs. As light verbs, they mean 'cause', 'let' and 'try', respectively (Annamalai, 2013).

As noted in the existing literature (Annamalai, 2013; Steever, 2005; Lehmann, 1993; Rajendran, 2004), Tamil is well known for a diverse type of Verb-Verb (V-V) and Noun-Verb (N-V) constructions.

- verb+verb constructions involve a series of auxiliary or light verbs that are added periphrastically to the first verb, either in the form of a verbal participle or an infinitive. Muthuchchanmugan (2005) shows that up to four verbal units can follow the main verb (also referred to as the lexical headword) in Tamil. However, as he claims, whether all of these are auxiliaries is debatable. In V-V constructions, the terminal verbal unit is the final item in a sequence. The preceding verbal units can be in either an adverbial or infinitival form. The terminal verbal unit is the item that carries all the functional information, such as tense, person, number, and gender. The V-V sequences are used to express a range of semantic information. This includes cross-linguistically well-established categories such as the causative, passive, permissive, negation, aspectual information, and mood and modality, including obligation vs. possibility. The literature also describes definitive and conclusive meanings, the expression of irritation, carelessness, augmentation, prediction and intention (Paramasivam, 2011; Muthuchchanmugan, 2005). For instance, Example (48) shows construction with definite conclusive meaning, where the *muṭi* 'finish' is the head verb that gives the meaning of complete and the auxiliary verb *viṭu* 'leave' (in the example, *viṭu* has become *viṭ* due to the phonological transformation) marks the definite completion.

- noun+verb constructions where the noun function as the head and the verb as a light verb. (49) is an example for such a construction where *kaitu* 'arrest' is the nominal head and *cey* 'do' is a light verb.

(48)
| அமைச்சர் | கூட்டத்தை | முடித்துவிட்டார் |
|---|---|---|
| amaiccar | kūṭṭattai | muṭittu-viṭṭār |
| minister.NOM | meeting.ACC | conclude-definite.PAST.3SER |

'The Minister concluded the meeting.'

(49) 
| பொலீஸ் | கள்வனைக் | கைது | செய்தது |
|---|---|---|---|
| polīs | kaḷvaṉaik | kaitu | ceytatu |
| police | thief.ACC.SANDHI_K | arrest.NOM | do.PAST.3SN |

'The police arrested the thief.'

### 3.5.2 Light Verb Construction

Light Verbs (LV) are also complex predicates. They differ from main/full verbs in their syntactic distribution and lexical semantics. While main verbs can stand alone and predicate independently, light verbs depend on the existence of another predicative element in the clause. LVs are light in the sense that they do not carry the meaning of the corresponding full verb, yet they still contain lexical-semantic information (Butt, 2010). Unlike auxiliaries, they are not fully functional elements. The light verb forms a syntactically monocausal unit with the main predicational element. Following Butt (2010), we can analyse LVs as a separate syntactic category and differentiate them from both main verbs and auxiliaries in the language. Such structures are common in South Asian Languages (Butt and Lahiri, 2013), including Tamil (Annamalai, 2013).

Annamalai (2013) has analysed various V-V, Infinitive-V, N-V and Verbal Participle-V sequences that have been analysed as Serial Verb Constructions (SVC) and Complex Predicates (CP).

Example (50) illustrates a simple transitive verb வாங்கு (*vangu*) 'buy'. An example of an N+V structure is given in (49).

The same main verb used together with கொடு (*kodu*) 'give' in its light verb sense forms a CP in (51). The light verb 'give' contributes a beneficiary meaning to the predication and licenses the use of an additional beneficiary indirect object (OBJ-TH). The light verb as the terminal verbal unit carries functional information, which in this case has to do with tense, number and person.

(50)
| நான் | காரை | வாங்கினேன் |
|---|---|---|
| naan | carai | vanginen |
| I.NOM | car.ACC | buy.PAST.1S |

'I bought the car.'

(51)
| நான் | அவனுக்குக் | காரை | வாங்கிக்கொடுத்தேன் |
|---|---|---|---|
| naan | avanukku-k | carai | vangikkoduththen |
| I.NOM | he.DAT-SAN | car.ACC | buy.VP.SAN.give.PAST.1S |

'I bought him a car.'

Example in (52) shows an alternative version of (51), in which the two parts of the complex predication are realised together. The *Sandhi* [k] is also triggered on the main verb வாங்கு (vangu) 'buy' just as in the single word realisation in (51), thus further consolidating the monocausal function of the whole construct.

(52)

| நான் | அவனுக்குக் | காரை | வாங்கிக் | கொடுத்தேன் |
|---|---|---|---|---|
| naan | avanukku-k | kar-ai | vangi-k | koduththen |
| I.NOM | he.NOM-SAN | car-ACC | buy.VP.SAN | give.PAST.1S |

'I bought car for him.'

### 3.5.3 Causatives

Causative verbs indicate that one person or thing causes another to do something for another person. A causative in Tamil can be realised either morphologically or syntactically.

**1. The morphological realisation of causation in Tamil**

The causation in Tamil can be morphologically realised via three morphs: வி *vi*, பி *pi*, and ப்பி *ppi* occur before the tense maker in a verb (Steever, 2005). For instance, (53) shows how the causative marker வி *vi* is used to causative a verb. The choice of the causative morph depends on the last vowel of the verbal root and is thus phonologically conditioned.

(53)

வாங்குவித்தான் (vanguvittaan)

| வாங்கு | -வி | -த் | -த் | -ஆன் |
|---|---|---|---|---|
| vangu | -vi | -t | -t | -aan |
| buy | -CAUS | -SAN | -PAST | -3SMR |

'he made somebody buy (something)'

**2. The syntactical realisation of causation in Tamil**

The syntactical realisation of causation in Tamil can also be realised by adding one of the following verbs after the infinitive form of the main verb: செய் *sei* 'do', வை *vai* 'put', பண்ணு *pannu* 'do'. In this case, these verbs do not predicate as full verbs and have the character of light verbs, expressing the non-referential meaning 'make', as shown in (54).

(54)

| அவனை | ஒரு | கார் | வாங்கச் | செய்தேன் |
|---|---|---|---|---|
| avanai | oru | car | vanga-c | seithen |
| he.ACC | a | car.NOM | buy.INF.SAN | make.PAST.3SM |

'(I) made him buy a car.'

Double causatives also exist in Tamil, where the causative form of a derivational verb can take an additional causative morphological marker to indicate double causation. This will be explored further in the next version of this research communication.

### 3.5.4 Passives in Tamil

Passive constructions in Tamil are realised via a V-V construction, and the verb படு *paṭu* 'be touched/be experienced/sleep' is used to passivise constructions, where *padu* functions as an auxiliary verb.

Together with an infinitive form of the main verb, it gives the meaning of 'be subjected to'. For instance, (55) is a passive construction where *paṭu* is added to the infinitival form of *vāṅku* to passivise the construction. Further, agent '*rām*' become an instrumental oblique and object '*car*' has become nominative subject.

(55) ராமால் கார் வாங்கப்பட்டது
rāmāl kār vāṅkappaṭṭatu
ram.INST car.NOM buy.INF.PASS.PAST.3SN
'A car was bought by Ram.'

(56) ராம் அவனுக்கு ஒரு கார் வாங்கிக் கொடுத்தான்
ram avanukku oru car vangki-k koduththaan
ram.NOM he.DAT a car.NOM buy.VP.SAN give.VP.PAST.3SM
'Ram bought a car for him.'

(57) ராமால் அவனுக்கு ஒரு கார் வாங்கிக் கொடுக்கப்பட்டது
ramaal avanukku oru car vangki-k kodukkappaddathu
ram.INST he.DAT a car.NOM buy.VP.SAN give.INF.SAN.PASS
'A car was bought for him by Ram.'

Further, V-V and N-V constructions can also be passivised. In such constructions, passiviser will be suffixed to the terminal auxiliary or light verb. For instance, consider (56) and its passive version in (57). As in causative CP constructions, passive constructions can also be written as two separate words, as in example (57), or like one token - வாங்கிக்கொடுக்கப்பட்டது *vāṅkikkoṭukkappaṭṭatu*. However, in the corpus, it was found that the passiviser verb *paṭu* is always written together as one token with the light verb as in (57) or the main verb.

### 3.5.5 Serial Verb Construction

Serial Verb Constructions (SVC) are common in Tamil. Unlike complex predicates, these constructions do not satisfy the constraint of co-predication and monoclausality, following the properties of SVCs outlined in Butt (1995), which differentiate SVC from other verbal predicates. (58) is an example of an SVC, where both *veddi* 'cut' and *vilttinaar* 'made-fall' share the same set of arguments and both of them are not bleached in their meaning. Anyway, SVCs need to be investigated in much more depth.

(58) விவசாயி மரத்தை வெட்டி வீழ்த்தினார்
vivasayi marattai veddi vilttinaar
farmer tree.NOM cut.VPART made-fall.PAST.3SER
'The farmer cut the tree down.'

### 3.5.6 Copula Construction

(59)  
| குமார் | வக்கீல் |
|---|---|
| kumār | vakkīl |
| Kumar.NOM | lawyer.NOM |

'Kumar is a lawyer.'

(59) is an example of a copular construction with a nominal predicate marked with the nominative case. The copula verb *āku* '*lit.* to become' optionally occurs in nominal predicates. *āku* can be used to identify the predicate when the subject and the predicate are in the nominative case. For instance, we can conjoin *āku* only with *vakkīl* to form a meaningful sentence in (59). The negative copula verb *illai* appears obligatorily to express constitution negation as in Example (60).

(60)  
| குமார் | வக்கீல் | இல்லை |
|---|---|---|
| kumār | vakkīl | illai |
| Kumar.NOM | lawyer.NOM | not |

'Kumar is not a lawyer.'

When the verb *illai* 'not' occurs as an existential negation, it is considered as the predicate as in example (61). There are instances where the nominal predicate is dative-case marked to express benefaction, as in example (62), and the copula is absent. In this example *āku* can only be suffixed to *kumārukku*, meaningfully; therefore, *kumārukku* is the predicate.

(61)  
| குமார் | வீட்டில் | இல்லை |
|---|---|---|
| kumār | vīṭṭ-il | illai |
| Kumar.NOM | home.LOC | not |

'Kumar is not at home.'

(62)  
| இந்த | பரிசு | குமாருக்கு |
|---|---|---|
| inta | paricu | kumārukku |
| this | gift.NOM | Kumar.DAT |

'This gift is for Kumar'

## 3.6 Clitics

Clitic is a word but cannot function independently because of its dependence on an adjoining word. Clitics are very common in Tamil. According to the Universal Dependencies guidelines, clitics should be separated from the main word to mark the corresponding syntactic functions in the dependency structure. The only clitics found in the corpus data are: *-ā, -tāṉ,* and *-um,* which mark an interrogative structure, the emphatic, and inclusiveness, respectively. For instance, examples (64)- (66) shows how interrogative marker

*-ā* is used along with the subject, the oblique noun, and the predicate to make different questions.

ஏ *ē* is another clitic mentioned in the literature for marking emphasis. This will be explored further in future versions of the communication.

(63) அவன் கொழும்புக்குப் போனான்
avan koḻumpukkup pōnān
he.NOM colombo.DAT.SANDHI_P go.PAST.3SM
'He went to Colombo.'

(64) அவனா கொழும்புக்குப் போனான்
avanā koḻumpukkup pōnān
he-who colombo.DAT.SANDHI-P go.PAST.3SM
'Is he who went to Colombo?'

(65) அவன் கொழும்புக்கா போனான்
avan koḻumpukkā pōnān
he.NOM colombo.DAT.-where go.PAST.3SM
'Is it to Colombo he went?'

(66) அவன் கொழும்புக்குப் போனானா
avan koḻumpukkup pōnānā
he.NOM colombo.DAT.SANDHI_P go.PAST.3SM-did he
'Did he go to Colombo?'

## 3.7 Coordination

Conjunctions in Tamil can be marked with the use of the clitic *-um* or the token மற்றும் *maṟṟum*. While the conjoined construction using *maṟṟum* is straightforward, *-um* is ambiguous in many cases, as it has at least eight different functions in syntax such as in-completion, superiority, doubt, negation, completion, number, definiteness, and that which is to come Senavaraiyar (1938).[18] The conjunction works in the same way for nouns, adjectives, adverbs and verbs. Apart from these markers, a coordinate structure can also be marked using sociative cases on the coordinating noun phrases. When conjoined noun phrases occur as the subject, subject-verbal predicate agreement will follow the precedence shown below, where rationality (rat) always takes precedence.

$NP_{1P+SG/PL+Rat} + NP_{1P+SG/PL+Rat} (+ NP_{3P+SG/PL+Irrat}) > VC_{1P+PL+Rat}$

$NP_{1P+SG/PL+Rat} + NP_{2P+SG/PL+Rat} (+ NP_{3P+SG/PL+Irrat}) > VC_{1P+PL+Rat}$

$NP_{2P+SG/PL+Rat} + NP_{2P+SG/PL+Rat} (+ NP_{3P+SG/PL+Irrat}) > VC_{2P+PL+Rat}$

$NP_{2P+SG/PL+Rat} + NP_{3P+SG/PL+Rat} (+ NP_{3P+SG/PL+Irrat}) > VC_{2P+PL+Rat}$

$NP_{3P+SG/PL+Rat} + NP_{3P+SG/PL+Rat} (+ NP_{3P+SG/PL+Irrat}) > VC_{3P+PL+Rat}$

$NP_{3P+SG/PL+Irrat} + NP_{3P+SG/PL+Irrat} > VC_{3P+PL+Irrat}$

---

[18] http://www.tamilvu.org/courses/degree/a021/a0213/html/a021331.htm

Disjunction is marked with the use of the clitic ஓ *ō* or the token அல்லது *allatu*. As in the case of coordination, this works in the same way irrespective of the disjunction of nouns, adjectives, adverbs, and verbs. Unlike the conjunction, it is unclear how the agreement between disjoined subjects and verbal predicates works. Since there were no such constructions in the corpus. This will be explored more in the future version of this communication.

## 3.8 Interrogatives

There are six types of interrogative structures in mentioned *naṉṉūl* (Thesikar, 1957). Their choice is based on the semantic nature of the question. Syntactically, all these questions can be constructed with the use of question particles or clitics *ō* and *ā*, or with the use of an in-situ question pronoun, as shown in (67) and (68), respectively.

(67) நீ    தோசையா    சோறா    சாப்பிடுகிறாய்
    nī    tōcaiyā    cōṟā    cāppiṭukiṟāy
    you   Dosa.*ā*   Rice.*ā*  eat.PAST.2S
    'Do you eat dosa or rice?'

(68) யார்   நாளைக்குக்    கொழும்புக்குப்   போகிறார்
    yār    nāḷaikkuk      koḻumpukkup      pōkiṟār
    who    tomorrow.DAT   Colombo.DAT      go.FUT.3SER
    'Who is going to Colombo tomorrow?'

## 3.9 Complex clauses in Tamil

Complex sentences are constructed with the use of one or more subordinate clauses. The embedded clause or the subordinate clause is a subconstituent of the main clause. Mostly, the subordinate clause precede the main verb of the main clause. However, since Tamil displays free word order, a subordinate clause can also follow the main verb of the main clause. This section briefly discusses the key types of complex sentence constructions found in the data that are dealt with.

## 3.10 Non-finite clauses

Tamil has several types of non-finite clauses, including infinitives, adjectivals, adverbials, and conditional clauses. The infinitive clause can mark the semantic interpretations of the verb, including perception and cognitive. The infinitives are marked with -*a* and mostly analysed as open clausal complement structures.

(69)

| நான் | குமாரை | கேட்க | வருகிறேன் |
|---|---|---|---|
| nān | kumārai | kēṭk-a | varukiṟēṉ |
| I.NOM | Kumar.ACC | ask.INF | come.PRES.1S |

'I come to ask Kumar.'

Adjectival clauses function like relative clauses, which will be discussed separately in Section 3.11. Adverbial clauses are constructed using the Verb+Adverbial-marker *-u* that modifies verbs, as shown in (70). These clauses are also used to construct the serial verb construction, as shown in 3.5.5.

(70)

| மாணவன் | புத்தகத்தைத் | திறந்து | பார்த்தான் |
|---|---|---|---|
| māṇavaṉ | puttakattait | tiṟantu | pārttāṉ |
| student.NOM | book.ACC.SANDHI_T | open.VPART | see.PAST.3SM |

'The student opened the book and looked.'

Conditional clauses are formed by Verb+Conditional-marker *-āl* that modify verbs, as shown in (71). This should not be confused with the *-āl* marker used to mark various aspects, including the instrumental case and the agent in passive constructions.

(71)

| கண்ணன் | வந்தால் | நான் | வர | மாட்டேன் |
|---|---|---|---|---|
| kaṇṇaṉ | vantāl | nāṉ | vara | māṭṭ.ēṉ |
| Kannan.NOM | come.CND | I.NOM | come.INF | will-not.1S |

'I will not come if Kannan comes.'

## 3.11 Relative Clauses in Tamil

Relative pronouns do not introduce relative clauses (RCs) in Tamil as in English. RCs are formed by adding an *-a* morph to the verb (72), which Butt et al. (2020) refer to as a relatives. The relative marker is null in the future participle form with *-um*, as shown in Example (73).

(72) [angu **nin-ṟ-a**] paiyan-ai naan paar-t-en
there **stand**-PAST-REL boy-ACC I.NOM.1S see-PAST-1S
'I saw the boy who stood there.'

(73) [angu **nirk-um-∅**] paiyan-ai naan paar-pp-en
there **stand**-FUT-REL boy-ACC I.NOM.1S see-FUT-1S
'I will see the boy who will stand there.'

The head noun of the RC in (73) is 'boy', with the relative clause directly preceding the head. One also finds RCs without a head noun in predicative contexts, as in (74). In this case, the verb in the RC instead carries the pronominal form *-atu*, apart from the relativiser *-a*. This *-atu* is form-identical with the indefinite pronoun *atu*, hence the English paraphrasing with the use of 'one'.

(74) [angu nin-ṛ-a-**athu**]           en thambi
    there stand-PAST-REL-PRON.3SN my brother
    'The **one** who stood there is my brother.'

Example (75) involves a full head noun 'boy' in the accusative as the matrix object, while (76) involves the substitution with the accusative pronoun *-avan* 'he' within the relative clause.

(75) [angu nin-ṛ-a]    **paiyan-ai** naan  paar-t-en
    there stand-PAST-REL boy-ACC   I.NOM see-PAST-1S
    'I saw the boy who stood there.'

(76) [angu nin-ṛ-a-**van**-ai]              naan  paar-t-en
    there stand-PAST-REL-PRON.3SM-ACC I.NOM see-PAST-1S
    'I saw the one (he) stood there.'

We have done an initial study on relative clauses and reported it in Butt et al. (2020). However, a more in-depth study on relative clauses is required, and will be covered in future revisions.

## 3.12  Complimentisers in Tamil

Tamil does not have complementisers of the *that*-type as in English. Rather, it uses a grammaticalised form of the verb *en* 'say', with the frozen past participle form *enṛu*, which has been analysed as a type of quotative (Amritavalli, 2013; Balusu, 2020). (77) displays the ambiguity which results between a quotative use and a complementiser function. (78) illustrates a pure complementiser reading.

(77) ravi         [naan  en nanban-ai santhi-tt-en] enṛu so-nn-an
    Ravi.3SM.NOM [Pron.1S my friend-ACC meet-PAST-1S] QUOT say-PAST-3SM
    'Ravi said that — "I met my friend".'
    'Ravi said that I met my friend.'

(78) ravi         [mazhai var-um        enṛu] ninai-tt-aan
    Ravi.3SM.NOM rain   come-FUT.3SN COMP think-PAST-3SM
    'Ravi thought that it will rain.'

Note that the matrix complementing verb can also take an accusative object as a co-referent for the complementiser clause as in (79). The pure complementiser form *enṛu* becomes *enṛa* by taking the relative marker *-a*. For further discussion on complementisers, see Butt et al. (2020).

(79) [avan pizhai sey-tt-aan     enṛ-a]    **unmaiy-ai** ram      nirupi-tt-aan
    he   mistake do-PAST-3SM COMP-REL truth-ACC Ram.NOM prove-PAST-3SM
    'Ram proved the truth that he made mistakes.'

## 3.13 Summary

This paper outlines the key and common morphology and syntactical constructions found in Tamil. The current constructions covered in this paper are drawn from a corpus collected for the purpose of constructing a rule-based grammar. This corpus will be expanded in the future, allowing for the extension of existing analyses and the inclusion of new ones. This is an evolving document.

## Acknowledgement

I extend my sincere thanks to the ZokoConnect/Herz Fellowship for their generous support during my stay at the University of Konstanz in the final quarter of 2023, which significantly facilitated the compilation of this article. My gratitude also goes to Professor Miriam Butt and Professor Gihan Dias for their insightful guidance/input during the paper's preparation. Additionally, I am grateful to all my collaborators whose diverse inputs were instrumental at various stages of my research.

# Appendix-I: Verb forms for நட *naṭa* 'walk'

| | | |
|---|---|---|
| நட | நடந்தவளை | நடப்பிக்கின்றவளால் |
| நடக்க | நடந்தவளைக் | நடப்பிக்கின்றவளுக்குக் |
| நடக்கக் | நடந்தவளைச் | நடப்பிக்கின்றவளுக்கு |
| நடக்கச் | நடந்தவளைத் | நடப்பிக்கின்றவளுக்குச் |
| நடக்கட்டும் | நடந்தவளைப் | நடப்பிக்கின்றவளுக்குத் |
| நடக்கத் | நடந்தவள் | நடப்பிக்கின்றவளுக்குப் |
| நடக்கப் | நடந்தவனால் | நடப்பிக்கின்றவளுடன் |
| நடக்கமுடியும் | நடந்தவனுக்குக் | நடப்பிக்கின்றவளை |
| நடக்கலாகாது | நடந்தவனுக்கு | நடப்பிக்கின்றவளைக் |
| நடக்கலாம் | நடந்தவனுக்குச் | நடப்பிக்கின்றவளைச் |
| நடக்கவில்லை | நடந்தவனுக்குத் | நடப்பிக்கின்றவளைத் |
| நடக்கவேண்டும் | நடந்தவனுக்குப் | நடப்பிக்கின்றவளைப் |
| நடக்காத | நடந்தவனுடன் | நடப்பிக்கின்றவள் |
| நடக்காதீர்கள் | நடந்தவனை | நடப்பிக்கின்றவனால் |
| நடக்காது | நடந்தவனைக் | நடப்பிக்கின்றவனுக்குக் |
| நடக்காதே | நடந்தவனைச் | நடப்பிக்கின்றவனுக்கு |
| நடக்காமல் | நடந்தவனைத் | நடப்பிக்கின்றவனுக்குச் |
| நடக்காவிட்டால் | நடந்தவனைப் | நடப்பிக்கின்றவனுக்குத் |
| நடக்கிறது | நடந்தவன் | நடப்பிக்கின்றவனுக்குப் |
| நடக்கிறது | நடந்தவார் | நடப்பிக்கின்றவனுடன் |
| நடக்கிறார் | நடந்தவார்கள் | நடப்பிக்கின்றவனை |
| நடக்கிறவரால் | நடந்தன | நடப்பிக்கின்றவனைக் |
| நடக்கிறவருக்குக் | நடந்தனள் | நடப்பிக்கின்றவனைச் |
| நடக்கிறவருக்கு | நடந்தனன் | நடப்பிக்கின்றவனைத் |
| நடக்கிறவருக்குச் | நடந்தாய் | நடப்பிக்கின்றவனைப் |
| நடக்கிறவருக்குத் | நடந்தார் | நடப்பிக்கின்றவன் |
| நடக்கிறவருக்குப் | நடந்தார்கள் | நடப்பிக்கின்றவார் |
| நடக்கிறவருடன் | நடந்தால் | நடப்பிக்கின்றவார்கள் |
| நடக்கிறவரை | நடந்தாள் | நடப்பிக்கின்றன |
| நடக்கிறவரைக் | நடந்தான் | நடப்பிக்கின்றனள் |
| நடக்கிறவரைச் | நடந்தீர் | நடப்பிக்கின்றன் |
| நடக்கிறவரைத் | நடந்தீர்கள் | நடப்பிக்கின்றாய் |
| நடக்கிறவரைப் | நடந்து | நடப்பிக்கின்றார் |
| | |  |

| நடக்கிறவர் | நடந்தும் | நடப்பிக்கின்றார்கள் |
| நடக்கிறவர்களால் | நடந்தேன் | நடப்பிக்கின்றாள் |
| நடக்கிறவர்களுக்குக் | நடந்தோம் | நடப்பிக்கின்றான் |
| நடக்கிறவர்களுக்கு | நடப்பது | நடப்பிக்கின்றீர் |
| நடக்கிறவர்களுக்குச் | நடப்பர் | நடப்பிக்கின்றீர்கள் |
| நடக்கிறவர்களுக்குத் | நடப்பவரால் | நடப்பிக்கின்றேன் |
| நடக்கிறவர்களுக்குப் | நடப்பவருக்குக் | நடப்பிக்கின்றோம் |
| நடக்கிறவர்களுடன் | நடப்பவருக்கு | நடப்பிக்கும் |
| நடக்கிறவர்களை | நடப்பவருக்குச் | நடப்பித்த |
| நடக்கிறவர்களைக் | நடப்பவருக்குத் | நடப்பித்தது |
| நடக்கிறவர்களைச் | நடப்பவருக்குப் | நடப்பித்தார் |
| நடக்கிறவர்களைத் | நடப்பவருடன் | நடப்பித்தவரால் |
| நடக்கிறவர்களைப் | நடப்பவரை | நடப்பித்தவருக்குக் |
| நடக்கிறவர்கள் | நடப்பவரைக் | நடப்பித்தவருக்கு |
| நடக்கிறவளால் | நடப்பவரைச் | நடப்பித்தவருக்குச் |
| நடக்கிறவளுக்குக் | நடப்பவரைத் | நடப்பித்தவருக்குத் |
| நடக்கிறவளுக்கு | நடப்பவரைப் | நடப்பித்தவருக்குப் |
| நடக்கிறவளுக்குச் | நடப்பவர் | நடப்பித்தவருடன் |
| நடக்கிறவளுக்குத் | நடப்பவர்களால் | நடப்பித்தவரை |
| நடக்கிறவளுக்குப் | நடப்பவர்களுக்குக் | நடப்பித்தவரைக் |
| நடக்கிறவளுடன் | நடப்பவர்களுக்கு | நடப்பித்தவரைச் |
| நடக்கிறவளை | நடப்பவர்களுக்குச் | நடப்பித்தவரைத் |
| நடக்கிறவளைக் | நடப்பவர்களுக்குத் | நடப்பித்தவரைப் |
| நடக்கிறவளைச் | நடப்பவர்களுக்குப் | நடப்பித்தவர் |
| நடக்கிறவளைத் | நடப்பவர்களுடன் | நடப்பித்தவர்களால் |
| நடக்கிறவளைப் | நடப்பவர்களை | நடப்பித்தவர்களுக்குக் |
| நடக்கிறவள் | நடப்பவர்களைக் | நடப்பித்தவர்களுக்கு |
| நடக்கிறவனால் | நடப்பவர்களைச் | நடப்பித்தவர்களுக்குச் |
| நடக்கிறவனுக்குக் | நடப்பவர்களைத் | நடப்பித்தவர்களுக்குத் |
| நடக்கிறவனுக்கு | நடப்பவர்களைப் | நடப்பித்தவர்களுக்குப் |
| நடக்கிறவனுக்குச் | நடப்பவர்கள் | நடப்பித்தவர்களுடன் |
| நடக்கிறவனுக்குத் | நடப்பவளால் | நடப்பித்தவர்களை |
| நடக்கிறவனுக்குப் | நடப்பவளுக்குக் | நடப்பித்தவர்களைக் |
| நடக்கிறவனுடன் | நடப்பவளுக்கு | நடப்பித்தவர்களைச் |
| நடக்கிறவனை | நடப்பவளுக்குச் | நடப்பித்தவர்களைத் |
| நடக்கிறவனைக் | நடப்பவளுக்குத் | நடப்பித்தவர்களைப் |
| நடக்கிறவனைச் | நடப்பவளுக்குப் | நடப்பித்தவர்கள் |
| நடக்கிறவனைத் | நடப்பவளுடன் | நடப்பித்தவளால் |
| நடக்கிறவனைப் | நடப்பவளை | நடப்பித்தவளுக்கக்க |
| நடக்கிறவன் | நடப்பவளைக் | நடப்பித்தவளுக்கு |
| நடக்கிறவார் | நடப்பவளைச் | நடப்பித்தவளுக்குச் |
| நடக்கிறவார்கள் | நடப்பவளைத் | நடப்பித்தவளுக்குத் |
| நடக்கிறன | நடப்பவளைப் | நடப்பித்தவளுக்குப் |
| நடக்கிறனள் | நடப்பவள் | நடப்பித்தவளுடன் |
| நடக்கிறனன் | நடப்பவனால் | நடப்பித்தவளை |
| நடக்கிறன் | நடப்பவனுக்குக் | நடப்பித்தவளைக் |
| நடக்கிறாய் | நடப்பவனுக்கு | நடப்பித்தவளைச் |
| நடக்கிறார் | நடப்பவனுக்குச் | நடப்பித்தவளைத் |
| நடக்கிறார்கள் | நடப்பவனுக்குத் | நடப்பித்தவளைப் |
| நடக்கிறார்கள் | நடப்பவனுக்குப் | நடப்பித்தவள் |
| நடக்கிறாள் | நடப்பவனுடன் | நடப்பித்தவனால் |
| நடக்கிறான் | நடப்பவனை | நடப்பித்தவனுக்குக் |
| நடக்கிறீர் | நடப்பவனைக் | நடப்பித்தவனுக்கு |
| நடக்கிறீர்கள் | நடப்பவனைச் | நடப்பித்தவனுக்குச் |
| நடக்கிறேன் | நடப்பவனைத் | நடப்பித்தவனுக்குத் |
| நடக்கிறோம் | நடப்பவனைப் | நடப்பித்தவனுக்குப் |
| நடக்கின்ற | நடப்பவன் | நடப்பித்தவனுடன் |
| நடக்கின்றது | நடப்பவார் | நடப்பித்தவனை |
| நடக்கின்றது | நடப்பவார்கள் | நடப்பித்தவனைக் |
| நடக்கின்றார் | நடப்பன் | நடப்பித்தவனைச் |
| நடக்கின்றவரால் | நடப்பாய் | நடப்பித்தவனைத் |
| நடக்கின்றவருக்குக் | நடப்பார் | நடப்பித்தவனைப் |
| நடக்கின்றவருக்கு | நடப்பார்கள் | நடப்பித்தவன் |
| நடக்கின்றவருக்குச் | நடப்பார்கள் | நடப்பித்தவார் |
| நடக்கின்றவருக்குத் | நடப்பாள் | நடப்பித்தவார்கள் |
| நடக்கின்றவருக்குப் | நடப்பான் | நடப்பித்தன |
| நடக்கின்றவருடன் | நடப்பிக்கிற | நடப்பித்தனள் |
| நடக்கின்றவரை | நடப்பிக்கிறது | நடப்பித்தனன் |
| நடக்கின்றவரைக் | நடப்பிக்கிறது | நடப்பித்தாய் |
| நடக்கின்றவரைச் | நடப்பிக்கிறர் | நடப்பித்தார் |
| நடக்கின்றவரைத் | நடப்பிக்கிறவரால் | நடப்பித்தார்கள் |
| நடக்கின்றவரைப் | நடப்பிக்கிறவருக்குக் | நடப்பித்தார்கள் |
| நடக்கின்றவர் | நடப்பிக்கிறவருக்கு | நடப்பித்தாள் |
| நடக்கின்றவர்களால் | நடப்பிக்கிறவருக்குச் | நடப்பித்தான் |
| நடக்கின்றவர்களுக்குக் | நடப்பிக்கிறவருக்குத் | நடப்பித்தீர் |
| நடக்கின்றவர்களுக்கு | நடப்பிக்கிறவருக்குப் | நடப்பித்தீர்கள் |
| நடக்கின்றவர்களுக்குச் | நடப்பிக்கிறவருடன் | நடப்பித்தேன் |
| நடக்கின்றவர்களுக்குத் | நடப்பிக்கிறவரை | நடப்பித்தோம் |
| நடக்கின்றவர்களுக்குப் | நடப்பிக்கிறவரைக் | நடப்பிப்பது |
| நடக்கின்றவர்களுடன் | நடப்பிக்கிறவரைச் | நடப்பிப்பர் |
| நடக்கின்றவர்களை | நடப்பிக்கிறவரைத் | நடப்பிப்பவரால் |
| நடக்கின்றவர்களைக் | நடப்பிக்கிறவரைப் | நடப்பிப்பவருக்குக் |
| நடக்கின்றவர்களைச் | நடப்பிக்கிறவர் | நடப்பிப்பவருக்கு |
| நடக்கின்றவர்களைத் | நடப்பிக்கிறவர்களால் | நடப்பிப்பவருக்குச் |
| நடக்கின்றவர்களைப் | நடப்பிக்கிறவர்களுக்குக் | நடப்பிப்பவருக்குத் |
| நடக்கின்றவர்கள் | நடப்பிக்கிறவர்களுக்கு | நடப்பிப்பவருக்குப் |
| நடக்கின்றவளால் | நடப்பிக்கிறவர்களுக்குச் | நடப்பிப்பவருடன் |
| நடக்கின்றவளுக்குக் | நடப்பிக்கிறவர்களுக்குத் | நடப்பிப்பவரை |
| நடக்கின்றவளுக்கு | நடப்பிக்கிறவர்களுக்குப் | நடப்பிப்பவரைக் |
| நடக்கின்றவளுக்குச் | நடப்பிக்கிறவர்களுடன் | நடப்பிப்பவரைச் |
| நடக்கின்றவளுக்குத் | நடப்பிக்கிறவர்களை | நடப்பிப்பவரைத் |
| நடக்கின்றவளுக்குப் | நடப்பிக்கிறவர்களைக் | நடப்பிப்பவரைப் |
| நடக்கின்றவளுடன் | நடப்பிக்கிறவர்களைச் | நடப்பிப்பவர் |
| நடக்கின்றவளை | நடப்பிக்கிறவர்களைத் | நடப்பிப்பவர்களால் |
| நடக்கின்றவளைக் | நடப்பிக்கிறவர்களைப் | நடப்பிப்பவர்களுக்குக் |
| நடக்கின்றவளைச் | நடப்பிக்கிறவர்கள் | நடப்பிப்பவர்களுக்கு |
| நடக்கின்றவளைத் | நடப்பிக்கிறவளால் | நடப்பிப்பவர்களுக்குச் |
| நடக்கின்றவளைப் | நடப்பிக்கிறவளுக்குக் | நடப்பிப்பவர்களுக்குத் |
| நடக்கின்றவள் | நடப்பிக்கிறவளுக்கு | நடப்பிப்பவர்களுக்குப் |
| நடக்கின்றவனால் | நடப்பிக்கிறவளுக்குச் | நடப்பிப்பவர்களுடன் |
| நடக்கின்றவனுக்குக் | நடப்பிக்கிறவளுக்குத் | நடப்பிப்பவர்களை |
| நடக்கின்றவனுக்கு | நடப்பிக்கிறவளுக்குப் | நடப்பிப்பவர்களைக் |
| நடக்கின்றவனுக்குச் | நடப்பிக்கிறவளுடன் | நடப்பிப்பவர்களைச் |
| நடக்கின்றவனுக்குத் | நடப்பிக்கிறவளை | நடப்பிப்பவர்களைத் |
| நடக்கின்றவனுக்குப் | நடப்பிக்கிறவளைக் | நடப்பிப்பவர்களைப் |
| நடக்கின்றவனுடன் | நடப்பிக்கிறவளைச் | நடப்பிப்பவர்கள் |
| நடக்கின்றவனை | நடப்பிக்கிறவளைத் | நடப்பிப்பவளால் |
| நடக்கின்றவனைக் | நடப்பிக்கிறவளைப் | நடப்பிப்பவளுக்குக் |
| நடக்கின்றவனைச் | நடப்பிக்கிறவள் | நடப்பிப்பவளுக்கு |
| நடக்கின்றவனைத் | நடப்பிக்கிறவனால் | நடப்பிப்பவளுக்குச் |
| நடக்கின்றவனைப் | நடப்பிக்கிறவனுக்குக் | நடப்பிப்பவளுக்குத் |
| நடக்கின்றவன் | நடப்பிக்கிறவனுக்கு | நடப்பிப்பவளுக்குப் |
| நடக்கின்றவார் | நடப்பிக்கிறவனுக்குச் | நடப்பிப்பவளுடன் |
| நடக்கின்றவர்கள் | நடப்பிக்கிறவனுக்குத் | நடப்பிப்பவளை |
| நடக்கின்றன | நடப்பிக்கிறவனுக்குப் | நடப்பிப்பவளைக் |
| நடக்கின்றனள் | நடப்பிக்கிறவனுடன் | நடப்பிப்பவளைச் |
| நடக்கின்றனன் | நடப்பிக்கிறவனை | நடப்பிப்பவளைத் |
| நடக்கின்றன் | நடப்பிக்கிறவனைக் | நடப்பிப்பவளைப் |
| நடக்கின்றாய் | நடப்பிக்கிறவனைச் | நடப்பிப்பவள் |
| நடக்கின்றார் | நடப்பிக்கிறவனைத் | நடப்பிப்பவனால் |
| நடக்கின்றார்கள் | நடப்பிக்கிறவனைப் | நடப்பிப்பவனுக்குக் |
| நடக்கின்றார்கள் | நடப்பிக்கிறவன் | நடப்பிப்பவனுக்கு |
| நடக்கின்றாள் | நடப்பிக்கிறவார் | நடப்பிப்பவனுக்குச் |
| நடக்கின்றான் | நடப்பிக்கிறவார்கள் | நடப்பிப்பவனுக்குத் |
| நடக்கின்றீர் | நடப்பிக்கிறன | நடப்பிப்பவனுக்குப் |
| நடக்கின்றீர்கள் | நடப்பிக்கிறனள் | நடப்பிப்பவனுடன் |
| நடக்கின்றேன் | நடப்பிக்கிறனன் | நடப்பிப்பவனை |
| நடக்கின்றோம் | நடப்பிக்கிறன் | நடப்பிப்பவனைக் |
| நடக்கும் | நடப்பிக்கிறாய் | நடப்பிப்பவனைச் |
| நடந்த | நடப்பிக்கிறார் | நடப்பிப்பவனைத் |
| நடந்தக்கக் | நடப்பிக்கிறார்கள் | நடப்பிப்பவனைப் |
| நடந்தது | நடப்பிக்கிறார்கள் | நடப்பிப்பவன் |
| நடந்தது | நடப்பிக்கிறாள் | நடப்பிப்பவார் |
| நடந்தார் | நடப்பிக்கிறான் | நடப்பிப்பவார்கள் |
| நடந்தவரால் | நடப்பிக்கிறீர் | நடப்பிப்பன் |
| நடந்தவருக்கக்க் | நடப்பிக்கிறீர்கள் | நடப்பிப்பாய் |
| நடந்தவருக்கு | நடப்பிக்கிறேன் | நடப்பிப்பார் |
| நடந்தவருக்குச் | நடப்பிக்கிறோம் | நடப்பிப்பார்கள் |
| நடந்தவருக்குத் | நடப்பிக்கின்ற | நடப்பிப்பாள் |
| நடந்தவருக்குப் | நடப்பிக்கின்றது | நடப்பிப்பான் |
| நடந்தவருடன் | நடப்பிக்கின்றர் | நடப்பிப்பீர் |
| நடந்தவரை | நடப்பிக்கின்றவரால் | நடப்பிப்பீர்கள் |
| நடந்தவரைக் | நடப்பிக்கின்றவருக்குக் | நடப்பிப்பியும் |
| நடந்தவரைச் | நடப்பிக்கின்றவருக்கு | நடப்பிப்பேன் |
| நடந்தவரைத் | நடப்பிக்கின்றவருக்குச் | நடப்பிப்போம் |
| நடந்தவரைப் | நடப்பிக்கின்றவருக்குத் | நடப்பீர் |
| நடந்தவர் | நடப்பிக்கின்றவருக்குப் | நடப்பீர்கள் |
| நடந்தவர்களால் | நடப்பிக்கின்றவருடன் | நடப்பேன் |
| நடந்தவர்களுக்குக் | நடப்பிக்கின்றவரை | நடப்போம் |
| நடந்தவர்களுக்கு | நடப்பிக்கின்றவரைக் | |
| நடந்தவர்களுக்குச் | நடப்பிக்கின்றவரைச் | |
| நடந்தவர்களுக்குத் | நடப்பிக்கின்றவரைத் | |
| நடந்தவர்களுக்குப் | நடப்பிக்கின்றவரைப் | |
| நடந்தவர்களுடன் | நடப்பிக்கின்றவர் | |
| நடந்தவர்களை | நடப்பிக்கின்றவர்களால் | |
| நடந்தவர்களைக் | நடப்பிக்கின்றவர்களுக்குக் | |
| நடந்தவர்களைச் | நடப்பிக்கின்றவர்களுக்கு | |
| நடந்தவர்களைத் | நடப்பிக்கின்றவர்களுக்குச் | |
| நடந்தவர்களைப் | நடப்பிக்கின்றவர்களுக்குத் | |
| நடந்தவர்கள் | நடப்பிக்கின்றவர்களுக்குப் | |
| நடந்தவளால் | நடப்பிக்கின்றவர்களுடன் | |
| நடந்தவளுக்குக் | நடப்பிக்கின்றவர்களை | |
| நடந்தவளுக்கு | நடப்பிக்கின்றவர்களைக் | |
| நடந்தவளுக்குச் | நடப்பிக்கின்றவர்களைச் | |
| நடந்தவளுக்குத் | நடப்பிக்கின்றவர்களைத் | |
| நடந்தவளுக்குப் | நடப்பிக்கின்றவர்களைப் | |
| நடந்தவளுடன் | நடப்பிக்கின்றவர்கள் | |